\documentclass[10pt,twocolumn,letterpaper]{article}

\usepackage{cvpr}
\usepackage{times}
\usepackage{epsfig}
\usepackage{graphicx}
\usepackage{amsmath}
\usepackage{amssymb}

\usepackage{tabularx}
\usepackage{subfig}
\usepackage{comment}
\usepackage{amsmath}
\usepackage{lineno}
\usepackage{amssymb}
\usepackage{bbm}

\newcolumntype{C}{>{\centering\arraybackslash}X}

\usepackage[breaklinks=true,bookmarks=false]{hyperref}

\cvprfinalcopy 

\def\cvprPaperID{****} 

\ifcvprfinal\fi
\begin{document}

\title{Consistency Regularization with High-dimensional Non-adversarial Source-guided Perturbation for Unsupervised Domain Adaptation in Segmentation}

\author{
    Kaihong Wang\\
    Boston University\\
    {\tt\small kaiwkh@bu.edu}
    \and
    Chenhongyi Yang\\
    Boston University\\
    {\tt\small hongyi@bu.edu}
    \and
    Margrit Betke\\
    Boston University\\
    {\tt\small betke@bu.edu}
}


\maketitle

\begin{abstract}
Unsupervised domain adaptation for semantic segmentation has been intensively studied due to the low cost of the pixel-level annotation for synthetic data. The most common approaches try to generate images or features mimicking the distribution in the target domain while preserving the semantic contents in the source domain so that a model can be trained with annotations from the latter. However, such methods highly rely on an image translator or feature extractor trained in an elaborated mechanism including adversarial training, which brings in extra complexity and instability in the adaptation process. Furthermore, these methods mainly focus on taking advantage of the labeled source dataset, leaving the unlabeled target dataset not fully utilized. In this paper, we propose a bidirectional style-induced domain adaptation method, called BiSIDA, that employs consistency regularization to efficiently exploit information from the unlabeled target domain dataset, requiring only a simple neural style transfer model. BiSIDA aligns domains by not only transferring source images into the style of target images but also transferring target images into the style of source images to perform high-dimensional perturbation on the unlabeled target images, which is crucial to the success in applying consistency regularization in segmentation tasks. Extensive experiments show that our BiSIDA achieves new state-of-the-art on two commonly-used synthetic-to-real domain adaptation benchmarks: GTA5-to-CityScapes and SYNTHIA-to-CityScapes.
\end{abstract}

\section{Introduction}


Deep learning methods for semantic segmentation~\cite{ShelhamerLoDa15}, the problem of dividing the pixels in an image into mutually exclusive and collectively exhaustive sets of class-labeled regions,  have gained increasing attention.
Research progress
is hindered by the difficulty of creating large training datasets with accurate pixel-level annotations of these regions.  
As a consequence, the use of synthetic datasets has become popular because pixel-level ground truth annotations can be generated along with the images. Unfortunately, when deep models that were trained on synthetic data are used to segment real-world images, their performance is typically limited due to the domain gap between the training and testing data.  Domain adaptation methods seek to bridge the gap between the source domain training data and the target domain testing data.  We here focus on unsupervised domain adaptation (UDA), the problem of adapting a model that was trained with a labeled source domain dataset, by using an unlabeled target domain dataset and optimizing its performance on the target domain.



To perform domain alignment on a pixel-level or feature-level basis, existing methods~\cite{TsaiHuScSoYaCh18,HoffmanTzPaZhIsSaEfDa18,VuJaBuCoPe19,LuoZhGuYuYa19,LiYuVa19,ChoiKiKi19} typically use adversarial training~\cite{GoodfellowPoMiXuWaOzCoBe14}, and training with the aligned data is then supervised by a loss computed with the annotation of the source domain dataset.
However, the use of adversarial training typically comes with extra complexity and instability in training. 
Alternative approaches~\cite{ZouYuKuWw18,VuJaBuCoPe19,LiYuVa19,ChoiKiKi19}
seek to exploit information about the unlabeled target dataset by performing semi-supervised learning including entropy minimization~\cite{GrandvaletBe04}, pseudo-labeling~\cite{Lee2013} and consistency regularization. However, these approaches either just play an auxiliary role in the training process besides supervised learning, or fail to take full advantage of the target dataset. 
In this paper, we propose Bidirectional Style-induced Domain Adaptation (BiSIDA) that takes better advantage of the unlabeled dataset and optimizes the performance of a segmentation model on the target dataset. Our pipeline includes a supervised learning phase that provides supervision using annotations in the source dataset 
and an unsupervised phase for learning from the unlabeled target dataset without requiring its annotation. To perform domain adaptation, we construct a non-adversarial yet effective pre-trained style-induced image generator that translates images through style transfer. In the supervised learning phase, the style-induced image generator translates images with different styles to align the source domain to the direction of the target domain. In the unsupervised phase, it performs high-dimensional perturbations on target domain images with consistency regularization.  Consequently, the unlabeled target dataset is utilized efficiently and the domain gap is reduced effectively
from another direction at the same time
through a self-supervised approach.

Our model performs image translation from the source to the target domain using an image generator in the supervised phase similar to existing methods. However, in order to facilitate generalization, our model synthesizes images with semantic content from the source domain, and with a style that is defined by a 
continuous parameter that represents a "mix" of source and target domain styles, instead of transferring the style directly to the target domain.
Consequently, the stochasticity of the whole process facilitates not only the training on the original images 
but also the gradual adaptation towards the target domain. 
The resulting image is then sent along with its corresponding pixel-level annotation to compute a supervised cross-entropy loss to train the segmentation model. 

BiSIDA employs consistency regularization in the unsupervised phase to yield consistent predictions on randomly perturbed inputs without requiring their annotations.
We apply our style-induced image generator as an augmentation method and transfer each target domain image together with a number of randomly sampled source domain images, just as in the supervised phase, but in an opposite direction. A series of images with identical content but different styles from source domain images is generated. Given that supervised learning is performed on source images that are transferred with combined styles of source images and target images,
our model will be more adapted and more likely to produce correct predictions when target domain images are transferred towards the direction of the source domain images.
Meanwhile, our style-induced image generator provides a high-dimensional perturbation that keeps the semantic content as indicated in~\cite{FrenchAiLaMaFi2019} for consistency regularization in a computational affordable way.
To further improve the quality of predictions, the transferred images are passed through the self-ensemble of the trained segmentation models, which is the exponential moving average of itself, and gathered to get a pseudo-label for the unlabeled target domain image. 
The training of the segmentation model on the original target domain image augmented with only color space perturbations is guided by its pseudo-label. During the process, information and knowledge lied in the unlabeled target images can be learned through the consistency regularization framework and the model is finally adapted to the target domain. 



Combined with our supervised and unsupervised learning methods, we are able to utilize annotation from the labeled source dataset, exploit knowledge from the unlabeled target dataset and perform gradual adaptation between the source and the target domain from both sides. 
In conclusion, our key contributions include:
\begin{enumerate}
  \item
    A Bidirectional Style-induced Domain Adaptation (BiSIDA) framework that incorporates both target-guided supervised and source-guided unsupervised learning. We also show that domain adaptation is achievable in a bidirectional way through a continuous parameterization of the two domains, without requiring adversarial training;
  \item
    A non-adversarial style-induced image generator that performs a high-dimensional source-guided perturbation on target images for consistency regularization.
  \item
    Extensive experiments show that our BiSIDA achieves new state-of-the-art on two commonly-used synthetic-to-real domain adaptation benchmarks: GTA5-to-CityScapes and SYNTHIA-to-CityScapes.
\end{enumerate}

\section{Related Works}

\subsection{Image-to-image Translation}

Recent progress in image-to-image translation that transfers the style of an image while preserving its semantic content has inspired research in various related areas, including image synthesis and reducing domain discrepancy. Typical image-to-image translation approaches include CycleGAN~\cite{ZhuPaIsEf17} and DualGAN~\cite{YiZhTaGo17}, which keep cycle-consistency in adversarial training to preserve the semantic content of images when transferring the style of image.
UNIT~\cite{LiuBrKa17} and MUNIT~\cite{HuangLiBeKa18} address the problem by mapping images into a common latent content space 
{\em Neural style transfer} offers an alternative way to perform image-to-image translation~\cite{GatysEcBe16}, but its optimization process is computationally impractical. Several works~\cite{JohnsonAlLi16, LiWa16,UlyanovLeVeLe16,UlyanovVeLe17,DumoulinShKu17} proposed improvements,
but these methods are limited since the style to be transferred is either fixed or the number of styles is limited. 

\subsection{Semi-supervised Learning}
\begin{figure*}[t!]
\begin{center}
    \includegraphics[width=\linewidth]{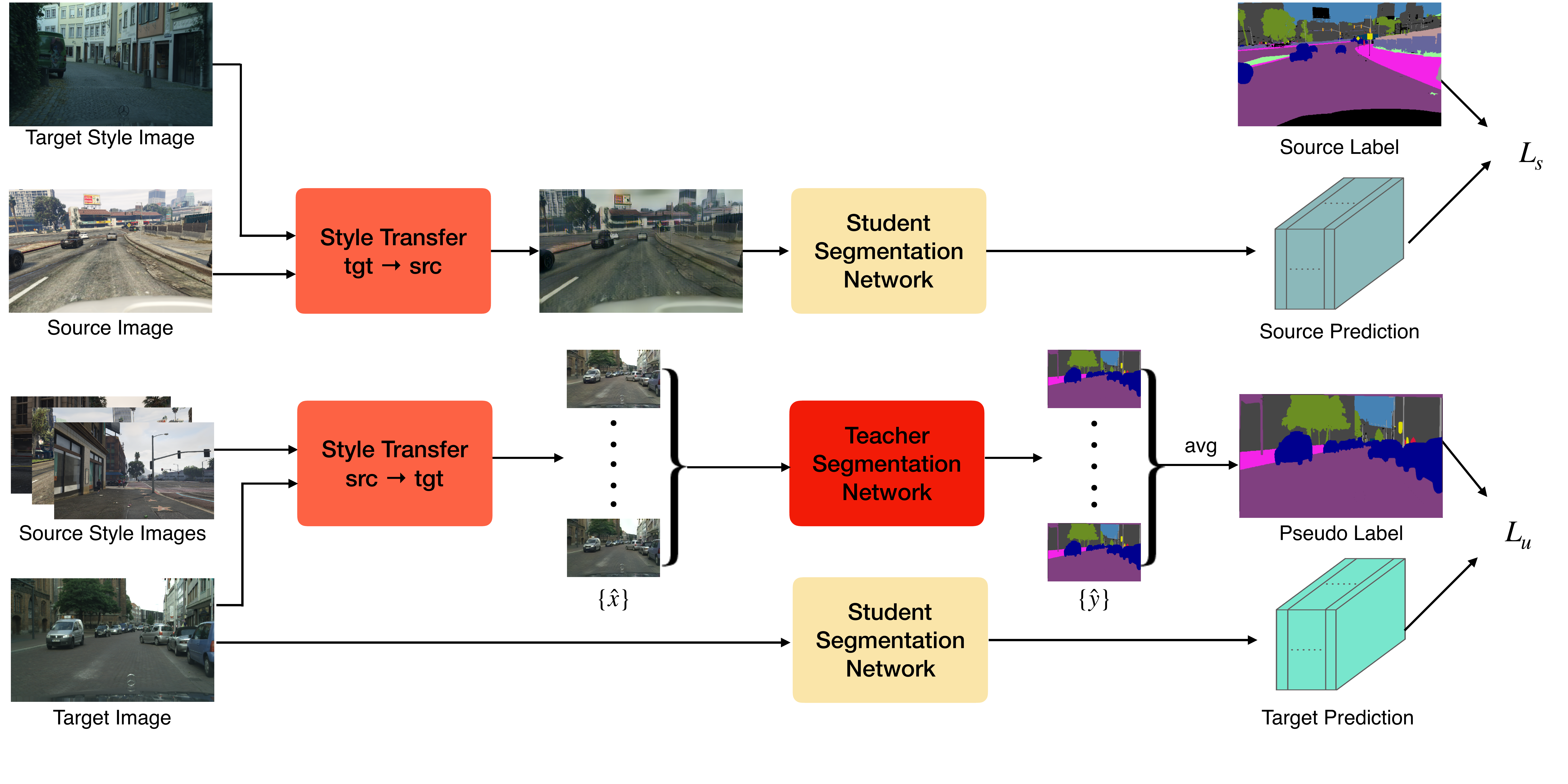}
    \vspace{-1cm}
\caption{The proposed training framework BiSIDA: It consists of two branches: 1) The supervised branch (top) augments a source-domain image through our style-induced image generator with the style of a target-domain image. A supervised segmentation loss~$L_{s}$ is computed with the corresponding annotation of the source image. 2) In the unsupervised learning branch (bottom), a target domain image and a series of source domain images are used to produce the corresponding translated images $\hat{x}$. Then each of these images will pass the through the teacher network to generate a set of probability maps $\hat{y}$. We computes an average of these maps to generate a pseudo-label, which is then used to compute $L_u$ for consistency regularization.}
\label{fig:pipeline}
\end{center}
\end{figure*}

When the gap between source and target domains 
becomes small,
the problem of unsupervised domain adaptation intriguingly degenerates to semi-supervised learning.  
Pseudo-labeling~\cite{Lee2013}, a commonly-used semi-supervised learning method, takes predictions on the unlabeled dataset with high confidence as one-hot labels guiding further training.
Entropy minimization~\cite{GrandvaletBe04} can be seen as a “soft assignment” of the pseudo-label on the unlabeled dataset. 
Recently, consistency regularization has gained attention due to its outstanding performance as  a semi-supervised learning method. The Mean-Teacher~\cite{TarvainenVa17} approach minimizes consistency loss on an unlabeled image between the output of a student network and the ensemble of itself, a teacher network. 
Fixmatch~\cite{sohnBeLiZhCaCuKuZhRa20} further outperforms Mean-Teacher by performing pseudo-labeling and consistency regularization between images with different degree of perturbations and achieves state-of-the-art
performance on several semi-supervised learning benchmarks. 

\subsection{UDA for Semantic Segmentation}

Current methods in UDA for segmentation can be categorized into adversarial and non-adversarial methods. "FCN in the wild""~\cite{HoffmanWaYuDa16} was the first to perform a segmentation task under UDA settings and align both global and local features between domains through adversarial training. Other works~\cite{HoffmanTzPaZhIsSaEfDa18,TsaiHuScSoYaCh18,VuJaBuCoPe19} tried to align features in one or multiple feature levels. 
The adversarial alignment process of each category between domains can be treated adaptively 
\cite{LuoZhGuYuYa19,WangYuWeFeXiHwHuSh20}. 
~\cite{ChoiKiKi19}
train an image translator in an adversarial way and take its output to perform consistency regularization. ~\cite{LiYuVa19} applied bidirectional learning in which an image translator and a segmentation model guide each other’s training in a mutual way. Pseudo-labeling is also performed to enhance performance.

Non-adversarial methods include 
a variety of techniques.
Curriculum DA~\cite{ZhangDaGo17} and PyCDA~\cite{LianDuLvGo19}, for example, adopt the concept of curriculum learning and align label distribution over images, landmark superpixels, or regions. CBST~\cite{ZouYuKuWw18} utilizes self-training to exploit information from the target domain images. DCAN~\cite{WuHaLiUzGoLiDa18} applies channel-wise alignment to merge the domain gap from both pixel-level and feature-level. Recently, \cite{YangSo20} proposed to align pixel-level discrepancy by performing a Fourier transformation. Combined with entropy minimization, pseudo-labeling and model ensemble, 
their method achieves current state-of-the-art performance.

The work that maybe most resembles ours is by~\cite{ChoiKiKi19}. However, our methods does not rely on a strong image translator that needs to be trained in an adversarial way. Furthermore, our method of adopting consistency regularization is able to exploit information more efficiently from target images by virtue of our high-dimensional perturbation method.

\section{Background}



BiSIDA uses 
{\bf Adaptive Instance Normalization, or AdaIN}~\cite{HuangBe17}, which consists of a encoder extracting a feature map from a given input image and a decoder that upsamples a feature map back to the original size of the input size. 
Given a content image $c$ and a style image $s$ from another domain, an image that mimic the style of $s$ while pertaining the content of $c$ will be synthesized. Practically, the encoder is taken from the first few layer of a pretrained fixed VGG-19~\cite{SimonyanZi15} while the decoder mirrors the architecture of the encoder and is trained as as proposed in \cite{HuangBe17}.
Formally, the feature map of a content image~$c$ and a style image~$s$ through an encoder $f$ can be represented as $t^c=f(c)$ and $t^s=f(s)$.
We can normalize the mean and the standard deviation for each channel
of $t^c$ and $t^s$ and produce the target feature maps $\hat{t}$:
\begin{linenomath}
\begin{align}
\hat{t}={\rm AdaIN}(t^c,t^s) &= \sigma(t^s)\left(\frac{t^c-\mu(t^c)}{\sigma(t^c)}\right) + \mu(t^s),
\end{align}
\end{linenomath}
where $\mu(t^*)$ and $\sigma(t^*)$ are the mean and variance of the feature maps.


A typical problem of training a model with pseudo-labels is the instability in the process caused by the uncertain quality of the pseudo-label.  It may lead to oscillation in predictions or bias to some easier classes. To stabilize the generation of pseudo-labels, we employ {\bf self-ensembling}~\cite{TarvainenVa17} which consists of a segmentation network as student network $F^s$ and a teacher network $F^t$ with the same architecture. The teacher network is essentially the temporal ensemble of the student network so that a radical change in the weight of the teacher network can be alleviated and more informed prediction can be made. The weight of the teacher network $F^t$ at the $i$th iteration $\theta_i^t$ is updated as the exponential moving average of the weight $\theta_i^s$ of the student network $F^s$,
$\theta_i^t=\eta\theta_{i-1}^t+(1-\eta)\theta_i^s$
given an exponential moving average decay $\eta$.


\section{Method}



In the UDA setting, the dataset from the source domain $\mathcal{S}$ includes images denoted by ${X^\mathcal{S}}$ with their corresponding pixel-level annotations denoted by ${Y^\mathcal{S}}$, and the dataset from the target domain $\mathcal{T}$ contains images represented by $X^\mathcal{T}$ without annotation. The task is to optimize a segmentation model using source dataset $\{(x_i^\mathcal{S}, y_i^\mathcal{S})\}_{(i=1,...,N^S)}$ and target images $\{(x_i^\mathcal{T})\}_{(i=1,...,N^T)}$ with $\mathcal{C}$ common categories. The student network is denoted by $F^s$, the teacher network by $F^t$.
The architecture of our model is shown in Figure~\ref{fig:pipeline}.

\subsection{Continuous Style-induced Image Generator}

To better utilize AdaIN to perform image augmentation in our framework, we control its output using content-style trade-off. Once the target feature map $\hat{t}$ is obtained, we can synthesize an image with the combined style with the style of a source and a target image controlled by a content-style trade-off parameter $\alpha$ varying from 0 to 1 through our image generator $G$:
\begin{linenomath}
\begin{align}
G(c,s,\alpha)=g(\alpha \hat{t} + (1-\alpha) t^c ),
\end{align}
\end{linenomath}
when $\alpha=0$, the content image will be reconstructed with its own style kept, and when $\alpha=1$, the output image will be the combination of the style of the style image $s$ and the content of the content image $c$. Finally, we rectify the output by clipping it in the range of $\left[ 0,255\right]$.

\subsection{Target-guided Supervised Learning}
Given a source domain dataset $\{(x_i^\mathcal{S}, y_i^\mathcal{S})\}_{(i=1,...,N^S)}$ and a target domain dataset $\{x_t^i\}$, we at first perform a random color space perturbation $\mathcal{A}$ on a source domain image $x_s$ to get $\mathcal{A}(x_s)$ to enhance the randomness. Images with color space perturbation augmentation will then be passed through our style-induced image generator $G$ to perform style transfer as a stronger augmentation method using a target domain image $x_t$. In the process, a content-style trade-off parameter $\alpha$ is randomly sampled from an uniform distribution $\mathcal{U}(0,1)$ to control the style of the translated image $\hat{x}_s=G(\mathcal{A}(x_s),x_t,\alpha)$. The translation process will be enabled with probability of $p_{s\rightarrow t}$ due to the loss of resolution in the translation process so the segmentation model can also be trained on details in images. For the rest of the probability, we simply assign $\mathcal{A}(x_s)$ to $\hat{x}_s$. Finally, we compute the supervised loss $L_s$ through a cross entropy loss between the probability map $p_s=F^s(\hat{x}_s)$ and its pixel-level annotation $y_s$:

\begin{linenomath}
\begin{align}
 L_{s}=-\frac{1}{HW}\sum_{m=1}^{H\times W}\sum_{c=1}^{C}y_s^{mc}log(p_s^{mc})
\end{align}
\end{linenomath}

Augmented by a strong and directed augmentation method, our framework facilitate generalization of model on images with different styles and further enable the adaptation towards the direction of the target domain.

\subsection{Source-guided Unsupervised Learning}

To start with, we introduce the generation of the pseudo-label that guides the  self-learning on the target dataset. Given that our model is more adapted to the source domain where our supervised learning is performed, the quality of produced pseudo-label is generally higher. Consequently, pseudo-label will be computed from target images transferred to the direction of the appearance of the source domain in our framework. Similar to the supervised phase, we at first perform a random color space perturbation $\mathcal{A}$ on a target domain image $x_t$ to get $\mathcal{A}(x_t)$. Then we augment each augmented target image $\mathcal{A}(x_t)$ using $k$ randomly sampled source images $\{x_s^i\}_{i=1}^k$ as style images through our style-induced image generator $G$ with probability of $p_{t\rightarrow s}$ for the consideration of the loss of resolution, and $x_t$ will be transferred to a set of images $\{\hat{x}_t^i\}_{i=1}^k$ where $\hat{x}_t^i=G(\mathcal{A}(x_t),x_s^i,\alpha)$. Otherwise it will simply be assigned to $\{\hat{x}_t^i\}_{i=1}^k=\{\mathcal{A}(x_t)\}$ with $k=1$. With the stochastic sampling of $k$ source images, our augmentation method performed on the target images will be stronger while their semantic meanings can also be preserved. After the augmentation process, transformed images $\{\hat{x}_t\}_{i=1}^k$ will be passed through the teacher model $F^t$ individually to acquire more stable predictions $\hat{y}^i=F^t(\hat{x}_t^i)$. We then average these predictions to get the probability map $p_l$ for the pseudo-label $p_l=\frac{1}{k}\sum_{i=1}^k\hat{y}^i$. 
Before the generation of pseudo-label, we employ a sharpening function which is widely adopted in various semi-supervised learning problems~\cite{BerthelotCaGoPaOlRa19} to re-arrange the distribution of the probability map as follows, given temperature $T$:

\begin{linenomath}
\begin{align}
{\rm Sharpening}(p,T)_i := \frac{p_i^{\frac{1}{T}}}{\sum_{j=1}^{C}p_j^{\frac{1}{T}}}
\end{align}
\end{linenomath}

Finally we can acquire the pseudo-label $q_t$ as $q={\rm argmax}(p_l)$, which can be used to comput the loss of our model on the target images in a supervised manner. Concretely, we augment the same target image $x_t$ using the random color space augmentation $\mathcal{A}$ and pass it through the student network $F^s$ to get the probability map $p_t=F^s(\mathcal{A}(x_t))$. 

In practice, the imbalance and complexity among categories in training datasets will cause the model to bias to popular or easier categories, especially when they are trained in a semi-supervised manner that relies on pseudo-label. To address this problem, we employ a class-balanced reweighting mechanism which guide the unsupervised loss with a prior distribution of categories. We first compute the class prior distribution $d_c$ as the portion of number of pixels over all categories on the source training dataset. Then the reweighting factor $w$ for each class is computed as:
\begin{linenomath}
\begin{align}
\label{eqn:reweight}
w_c=\frac{1}{\lambda \: d_c^\gamma},
\end{align}
\end{linenomath}
where $\lambda$ and $\gamma$ are hyper-parameters. Thus, the final unsupervised loss is presented as:
\begin{linenomath}
\begin{align}
 L_{u}=-\frac{1}{HW}\sum_{m=1}^{H\times W}\mathbbm{1}({\rm max}(p_l^m)\geq\tau)\sum_{c=1}^{C}w_cq^{mc}log(p_t^{mc})
\end{align}
\end{linenomath}

\subsection{Optimization}
To summarize, our framework comprises a supervised learning process performed on the labeled source dataset as well as an unsupervised learning process performed on the unlabeled target dataset via consistency regularization and pseudo-labeling. As a result, we can compute the final loss $L$, given the weight of the unsupervised loss $\lambda_u$ in a multi-task learning manner, as follows:
\begin{linenomath}
\begin{align}
L=L_{s}+\lambda_u L_{u}.
\end{align}
\end{linenomath}
During the training process, the weight of the student network $F_s$ is updated toward the direction of the gradient computed via back-propagation of the loss $\ L$, while the weight of the teacher network is updated as the exponential moving average of the student network.



\section{Experiments}

\begin{figure*}[t!]
\begin{center}
    \includegraphics[width=\linewidth]{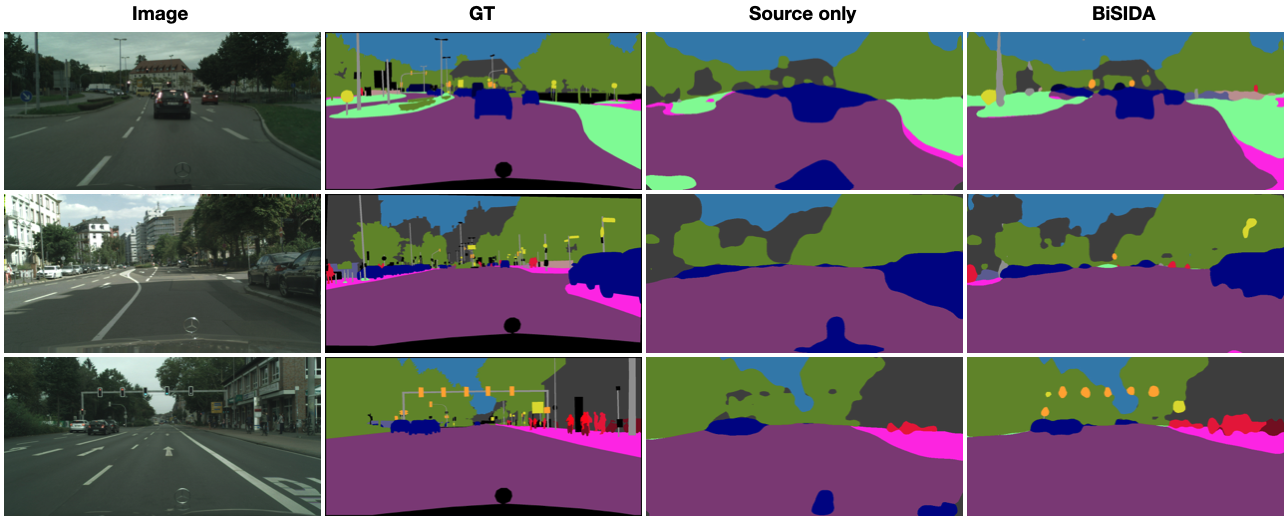}
\caption{Sample results (column "BiSIDA"). Target-domain testing images from the CityScapes dataset (column "Images") were segmented with a model trained by our BiSIDA framework on the GTA5 dataset through a FCN-8s with a VGG-16 backbone.  The ground truth segmentation ("GT") and results of a model trained only with source domain images ("Source only") are also shown.  Note that our method 
is capable of capturing rare and difficult categories, such as traffic lights and signs.
}
\label{fig:vis}
\end{center}
\end{figure*}

Extensive experiments are made on two commonly used synthetic-to-real segmentation benchmarks. Comparisons with other SOTA methods and ablation studies are presented to show the effectiveness of our BiSIDA framework.  We visualize some segmentation results in Figure~\ref{fig:vis}.


\begin{table*}[h!]
  \begin{center}
  \resizebox{\textwidth}{!}{
  \begin{tabular}{c|ccccccccccccccccccc|c}
    \hline
    Method&\rotatebox{60}{road}&\rotatebox{60}{sidewalk}&\rotatebox{60}{building}&\rotatebox{60}{wall}&\rotatebox{60}{fence}&\rotatebox{60}{pole}&\rotatebox{60}{light}&\rotatebox{60}{sign}&\rotatebox{60}{vegetation}&\rotatebox{60}{terrain}&\rotatebox{60}{sky}&\rotatebox{60}{person}&\rotatebox{60}{rider}&\rotatebox{60}{car}&\rotatebox{60}{truck}&\rotatebox{60}{bus}&\rotatebox{60}{train}&\rotatebox{60}{motocycle}&\rotatebox{60}{bicycle}&mIoU \\
    \hline
    \hline
    Curriculum~\cite{ZhangDaGo17}&74.9&22.0&71.7&6.0&11.9&8.4&16.3&11.1&75.7&13.3&66.5&38.0&9.3&55.2&18.8&18.9&0.0&16.8&16.6&29.0 \\
    CBST~\cite{ZouYuKuWw18}&66.7&26.8&73.7&14.8&9.5&28.3&25.9&10.1&75.5&15.7&51.6&47.2&6.2&71.9&3.7&2.2&5.4&18.9&32.4&30.9 \\
    AdaSeg~\cite{TsaiHuScSoYaCh18}&87.3&29.8&78.6&21.1&18.2&22.5&21.5&11.0&79.7&29.6&71.3&46.8&6.5&80.1&23.0&26.9&0.0&10.6&0.3&35.0 \\
    Cycada~\cite{HoffmanTzPaZhIsSaEfDa18}&85.2&37.2&76.5&21.8&15.0&23.8&22.9&21.5&80.5&31.3&60.7&50.5&9.0&76.9&17.1&28.2&4.5&9.8&0.0&35.4 \\
    AdvEnt~\cite{VuJaBuCoPe19}&86.9&28.7&78.7&28.5&25.2&17.1&20.3&10.9&80.0&26.4&70.2&47.1&8.4&81.5&26.0&17.2&18.9&11.7&1.6&36.1 \\
    DCAN~\cite{WuHaLiUzGoLiDa18}&82.3&26.7&77.4&23.7&20.5&20.4&30.3&15.9&80.9&25.4&69.5&52.6&11.1&79.6&24.9&21.2&1.3&17.0&6.7&36.2 \\
    CLAN~\cite{LuoZhGuYuYa19}&88.0&30.6&79.2&23.4&20.5&26.1&23.0&14.8&81.6&34.5&72.0&45.8&7.9&80.5&26.6&29.9&0.0&10.7&0.0&36.6 \\
    LSD~\cite{SankaranarayananBaJaLiCh18}&88.0&30.5&78.6&25.2&23.5&16.7&23.5&11.6&78.7&27.2&71.9&51.3&19.5&80.4&19.8&18.3&0.9&20.8&18.4&37.1 \\
    BDL~\cite{LiYuVa19}&89.2&40.9&81.2&29.1&19.2&14.2&29.0&19.6&83.7&35.9&80.7& {54.7}&23.3&82.7&25.8&28.0&2.3&25.7&19.9&41.3 \\
    FDA~\cite{YangSo20}&86.1&35.1&80.6&30.8&20.4&27.5&30.0&26.0&82.1&30.3&73.6&52.5&21.7&81.7&24.0&30.5&29.9&14.6&24.0&42.2 \\
    Stuff\&things~\cite{WangYuWeFeXiHwHuSh20}&88.1&35.8&83.1&25.8&23.9&29.2&28.8&28.6&83.0&36.7&82.3&53.7&22.8&82.3&26.4&38.6&0.0&19.6&17.1&42.4 \\
    TGCF-DA+SE~\cite{ChoiKiKi19}&90.2&51.5&81.1&15.0&10.7&37.5&35.2&28.9&84.1&32.7&75.9&62.7&19.9&82.6&22.9&28.3&0.0&23.0&25.4&42.5 \\
    \hline
    \textbf{Ours} &89.3&40.9&82.5&30.9&24.7&20.9&26.9&32.1&81.8&33.1&81.6&53.4&20.3&83.0&24.8&29.4&0.0&28.6&36.6&\textbf{43.2}\\
    \hline
  \end{tabular}
  }
    \end{center}
    \caption{Comparison of our model with other methods on the GTA5-to-CityScapes benchmark using models with VGG-16 as backbone. The ${\rm mIoU}$ represents the average of individual mIoUs among all 19 categories between GTA5 and CityScapes.}
  \label{tab:GTA_experiments}
\end{table*}
\begin{table*}[t!]
  \begin{center}
  \resizebox{\textwidth}{!}{
  \begin{tabular}{c|cccccccccccccccc|cc}
    \hline
    Method&\rotatebox{60}{road}&\rotatebox{60}{sidewalk}&\rotatebox{60}{building}&\rotatebox{60}{wall}&\rotatebox{60}{fence}&\rotatebox{60}{pole}&\rotatebox{60}{light}&\rotatebox{60}{sign}&\rotatebox{60}{vegetation}&\rotatebox{60}{sky}&\rotatebox{60}{person}&\rotatebox{60}{rider}&\rotatebox{60}{car}&\rotatebox{60}{bus}&\rotatebox{60}{motocycle}&\rotatebox{60}{bicycle}&mIoU&mIoU* \\
    \hline
    \hline
    Curriculum~\cite{ZhangDaGo17}&65.2&26.1&74.9&0.1&0.5&10.7&3.5&3.0&76.1&70.6&47.1&8.2&43.2&20.7&0.7&13.1&29.0&34.8 \\
    AdvEnt~\cite{VuJaBuCoPe19}&67.9&29.4&71.9&6.3&0.3&19.9&0.6&2.6&74.9&74.9&35.4&9.6&67.8&21.4&4.1&15.5&31.4&36.6 \\
    AdaSeg~\cite{TsaiHuScSoYaCh18}&78.9&29.2&75.5&-&-&-&0.1&4.8&72.6&76.7&43.4&8.8&71.1&16.0&3.6&8.4&-&37.6 \\
    CLAN~\cite{LuoZhGuYuYa19}&80.4&30.7&74.7&-&-&-&1.4&8.0&77.1&79.0&46.5&8.9&73.8&18.2&2.2&9.9&-&39.3 \\
    CBST~\cite{ZouYuKuWw18}&69.6&28.7&69.5&12.1&0.1&25.4& {11.9}&13.6&82.0& {81.9}&49.1&14.5&66.0&6.6&3.7&32.4&35.4&40.7 \\
    DCAN~\cite{WuHaLiUzGoLiDa18}&79.9&30.4&70.8&1.6& {0.6}&22.3&6.7&23.0&76.9&73.9&41.9&16.7&61.7&11.5&10.3&38.6&35.4&41.7 \\
    LSD~\cite{SankaranarayananBaJaLiCh18}&80.1&29.1&77.5&2.8&0.4&26.8&11.1&18.0&78.1&76.7&48.2&15.2&70.5&17.4&8.7&16.7&36.1&42.1 \\
    ROAD~\cite{ChenLiGo18}&77.7&30.0&77.5&9.6&0.3&25.8&10.3&15.6&77.6&79.8&44.5&16.6&67.8&14.5&7.0&23.8&36.2&41.7 \\
    GIO-Ada~\cite{ChenLiChGo19}&78.3&29.2&76.9&11.4&0.3&26.5&10.8&17.2&81.7& {81.9}&45.8&15.4&68.0&15.9&7.5&30.4&37.3&43.0 \\
    TGCF-DA+SE~\cite{ChoiKiKi19}& {90.1}& {48.6}& {80.7}&2.2&0.2& {27.2}&3.2&14.3& {82.1}&78.4&54.4&16.4& {82.5}&12.3&1.7&21.8&38.5&46.6 \\
    BDL~\cite{LiYuVa19}&72.0&30.3&74.5&0.1&0.3&24.6&10.2& {25.2}&80.5&80.0&54.7& {23.2}&72.7&24.0&7.5& {44.9}&39.0&46.1 \\
    FDA~\cite{YangSo20}&84.2&35.1&78.0&6.1&0.4&27.0&8.5&22.1&77.2&79.6& {55.5}&19.9&74.8&24.9&14.3&40.7&40.5&47.3 \\
    \hline
     \textbf{Ours} &87.4&42.4&79.0& {17.0}&0.1&23.9&2.8&22.9&82.0&80.4&51.1&19.1&76.7& {33.3}& {14.4}&41.2& \textbf{42.1}& \textbf{48.7}\\
    \hline
  \end{tabular}}
  \end{center}
   \caption{Comparison of our framework with other methods on SYNTHIA to CityScapes benchmark using models with VGG-16 as backbone. The ${\rm mIoU}$ represents the average of individual mIoUs among all 16 categories between SYNTHIA and CityScapes while the ${\rm mIoU^*}$ represents that among 13 common categories excluding wall, fence and pole.}
  \label{tab:Synthia_experiments}
\end{table*}
\subsection{Datasets}

We used two synthetic-to-real benchmarks, GTA5-to-Ci\-ty\-Scapes and SYNTHIA-to-CityScapes. The CityScapes dataset~\cite{CordtsOmRaReEnBeFrRoSc16} consists of images of real street scenes of spatial resolution of 2048$\times$1024 pixels. It includes 2,975 images for training, 500 images for validation, and 1,525 images for testing. In our experiments, we used the 500 validation images as a test set. The GTA5 dataset~\cite{RichterViRoKo16} includes 24,966 synthetic images with a resolution of 1914$\times$1052 pixels that are obtained from the video game GTA5 along with pixel-level annotations that share all 19 common categories of CityScapes. For the SYNTHIA dataset~\cite{RosSeMaVaLo16}, we used the SYNTHIA-RAND-CITYSCAPES subset, which contains 9,400 rendered images of size 1280$\times$760 and shares 16 common categories with the CityScapes dataset.

\subsection{Network Architecture}
 
\textbf{Image generator:} To keep our continuous style-induced image generator light-weighted and computationally affordable, we adopted the first several layers up to relu4\_1 of a fixed pre-trained VGG-19 network as the encoder in our experiments. For the decoder, we reversed the order of layers in the encoder and replaced the pooling layers by nearest up-sampling~\cite{HuangBe17}. 

\textbf{Segmentation network:} We chose FCN-8s~\cite{ShelhamerLoDa15} with a VGG16 backbone network, pre-trained with ImageNet. 

\subsection{Training Protocol}
The \textbf{continuous style-induced image generator} was trained using randomly-cropped $640\times320$ images, and a batch size of 4. The ADAM optimizer was used with a learning rate of $1\times10^{-5}$ and momentum of 0.9 and 0.999. To balance the reconstruction of the content image and the extraction from the style image, we optimized the generator loss in~\cite{HuangBe17} and with style weight $0.1$. 
The \textbf{segmentation model} was trained on images randomly cropped to $960\times480$ with batch size of 1. On the GTA5 dataset, we applied the ADAM optimizer with a learning rate of $1\times10^{-5}$, weight decay of $5\times10^{-4}$ and momentum of 0.9 and 0.999. For the SYNTHIA dataset, we adopted the SGD optimizer with a learning rate of $1\times10^{-5}$, momentum of 0.99 and and weight decay of $5\times10^{-4}$. We set the exponential moving average decay for the teacher model to 0.999 and the confidence threshold~$\tau$ in the pseudo-label generation process to 0.9. The probability of performing target-guided image translation $p_{s\rightarrow t}$ and  source-guided image translation $p_{t\rightarrow s}$ is 0.5. The unsupervised weight $\lambda_{u}$ and the sharpening temperature $T$ are set to $1$ and $0.25$, respectively.
Both models are trained on a NVIDIA Tesla V100 GPU.  

\subsection{Comparisons with SOTA Methods}
We first compare the performance of our BiSIDA on the GTA5-to-CityScapes benchmark with that of other methods using models with VGG-16 as backbone
(Table~\ref{tab:GTA_experiments}). Our results reveal that our method outperforms most competitive methods, especially TGCF-DA+SE, which employs adversarial training as augmentation and achieves state-of-the-art performance by $1.6\%$.

We present the performance of our and other methods on the SYNTHIA-to-CityScapes benchmark using two metrics (Table~\ref{tab:Synthia_experiments}). Due to the less realistic appearance and fewer training data, this task is more difficult than the previous one. However, our framework outperforms the current state-of-the-art method by a significant margin of $3.8\%$.

\subsection{Ablation Studies}
\textbf{Style-induced image translation and unsupervised learning:} We validate the effectiveness of our continuous style-induced image generator as well as our self-supervised learning modules through an ablation study, and explore how they contribute to achieve unsupervised domain adaptation. Results are presented in Table~\ref{tbl:ablation}. Since our continuous style-induced image generator is used in both the supervised and the unsupervised learning phase to perform a target-guided and a source-guided image translation, we conduct experiments on both of them respectively. Additionally, given that our self-supervised learning paradigm is based on the source-guided image translation, we deactivate the self-supervised learning when the source-guided image translation is suppressed in this experiment. As we can observe from the results, the target-guided and source-guided image translation improve the performance on both benchmarks when applied separately. It is also worth noting that the improvement brought by the target-guided image translation is slightly larger since the target domain images translated with styles from source domain cannot provide better self-guidance without having the source domain aligned to the intermediate continuous space. A more significant performance leap is shown when these two translations are performed simultaneously, especially on SYNTHIA-to-CityScapes 
benchmark where domain gap is larger, showing the advantage of our bidirectional style-induced image translation method.


\begin{table}[t]
  \begin{center}
    \begin{tabularx}{\columnwidth}{CCCC|C|C}
      \hline
      S2T & T2S & PL & SE & GTA & SYN \\
      \hline
               &            &              &            & 29.3 & 28.9 \\
    \checkmark &            &              &            & 34.7 & 32.0 \\
               &\checkmark  &              &            & 31.8 & 31.4 \\
    \checkmark & \checkmark &              &            & 35.1 & 40.2 \\
    \hline
    \checkmark & \checkmark &   \checkmark &            & 35.4 & 40.8 \\
    \checkmark & \checkmark &              & \checkmark & 39.4 & 41.8 \\
    \checkmark & \checkmark &   \checkmark & \checkmark & \textbf{43.2} & \textbf{42.1} \\
    \hline
    \end{tabularx}
  \end{center}
   \caption{Ablation study on the style-induced image translation and unsupervised modules on SYNTHIA to CityScapes benchmark. S2T stands for source domain to target domain image translation, T2S stands for target to source domain image translation, PL stands for pseudo-labeling and SE stands for self-ensembling. GTA represents the mIoU (16 classes) from GTA5 to CityScapes benchmark while SYN represents the mIoU from SYNTHIA to CityScapes} 
\label{tbl:ablation}
\end{table}

As for the modules in unsupervised learning phase, we explore the capability of pseudo-labeling and self-ensembling. When pseudo-labeling is disabled, we use the probability maps to compute the self-supervised loss and the problem will be transformed to entropy minimization. Also, the probability maps will be generated by the segmentation model itself if self-ensembling is disabled. From the results, we can find that both pseudo-labeling and self-ensembling contribute to similar degree of enhancement in the performance. Additionally, we may also observe that most of the improvement on GTA5-to-CityScapes comes from the application of self-supervised learning modules while that on SYNTHIA-to-CityScapes, on the other hand, comes from the style-induced image translation process. Based on such observation, we can infer that the challenge in the GTA5-to-CityScapes benchmark is to perform feature-level alignment while for SYNTHIA-to-CityScapes is to perform pixel-level alignment.
\begin{table}[t]
  \begin{center}
    \begin{tabularx}{\columnwidth}{CC|CC}
      \hline
      cAUG & T2S & mIoU & mIoU* \\
      \hline
                 &            & 32.2 & 38.6  \\
      \checkmark &            & 32.1 & 38.5 \\
                 & \checkmark & 41.8 & 48.3 \\
      \checkmark & \checkmark & \textbf{42.1} & \textbf{48.7}  \\
    \hline
    \end{tabularx}
  \end{center}
  \caption{Experiments on augmentation methods on SYNTHIA to CityScapes. cAUG represents color perturbation, T2S represents source-guided image translation performed on target domain. mIoU represents averaged mIoU over 16 classes and mIoU* represents that over 13 common classes.} 
\label{tbl:augment_target}
\end{table}


\textbf{Source-guided image translation:} In the previous experiment, the unsupervised learning was suppressed when source-guided image translation was not performed. To learn more about the effectiveness of the 
color-space perturbation and source-guided image translation 
performed on target images in the unsupervised learning phase, we conducted an ablation study on the SYNTHIA-to-CityScapes benchmark, where pixel-level alignment plays a more important role. We tested these two perturbation methods with all other settings fixed. From the results, shown in Table~\ref{tbl:augment_target}, we find that the introduction of source-guided image translation significantly improves performance by a large margin. 
On the other hand, the color space perturbation only helps when the source-guided image translation is applied since it enhances the stochasticity in the high-dimensional perturbation process. 
Otherwise, the color space perturbation is not a sufficiently strong  perturbation method for consistency regularization.

\begin{table}[t]
  \begin{center}
    \begin{tabularx}{\columnwidth}{C|CCCCC}
      \hline
      weight & 0.1 & 0.5 & 1.0 & 5.0 & 10.0 \\
      \hline
       mIoU  & 37.8 & 41.9 & \textbf{42.1} & 39.8 & 38.6  \\
       mIoU* & 44.3 & 48.1 & \textbf{48.7} & 46.3 & 45.2 \\
      \hline
    \end{tabularx}
  \end{center}
  \caption{Comparison with different unsupervised loss weigths $\lambda_{u}$. mIoU represents averaged mIoU over 16 classes and mIoU* represents that over 13 common classes.} 
\label{tbl:unsup-weight}
\end{table}

\subsection{Discussion}
\textbf{Unsupervised learning weight:} In our BiSIDA, the unsupervised loss weight $\lambda_{u}$ is a crucial hyperparameter to balance the focus of our model between the supervised learning on the labeled source dataset and the self-supervised learning on the unlabeled target dataset. To investigate the effect of using different unsupervised loss weights on our method, 
we conducted an experiment on the SYNTHIA-to-CityScapes benchmark with five different unsupervised loss weights
The results in Table~\ref{tbl:unsup-weight} reveal that when the weight is too small, 
the benefit of unsupervised learning is 
limited and 
consistency regularization cannot be performed effectively. When the weight is too large, the model fails to achieve satisfying performance. 
A reason may be that the model becomes bias prone and prefers
an easier category in the early stage of training. Our model reaches the peak of performance when the weight is set to $1$.
\begin{table}[t]
  \begin{center}
    \begin{tabularx}{\columnwidth}{C|CCCCC}
      \hline
      \# img & 1 & 2 & 4 & 6 & 8 \\
      \hline
       mIoU  & 41.0 & 41.4 & \textbf{42.1} & 41.8 & 42.0  \\
       mIoU* & 47.3 & 47.6 & \textbf{48.7} & 48.1 & 48.6 \\
      \hline
    \end{tabularx}
  \end{center}
  \caption{Numer of source images used to perturb a target image. mIoU represents averaged mIoU over 16 classes and mIoU* represents that over 13 common classes.} 
\label{tbl:kt}
\end{table}

\textbf{Number of style images used in source-guided image translation:} Since we gather the predictions over $k$ images translated from a target domain image with styles from $k$ different source domain style images, the number of images used in the image translation process is another important hyperparameter in our BiSIDA framework. We hereby conduct experiments on SYNTHIA-to-CityScapes benchmark with $k$ value of $1$, $2$, $4$, $6$ and $8$ respectively. The results are presented in Table~\ref{tbl:kt}. As we can see from the table, when the number of style images is smaller, the model cannot achieve a good performance since the stochasticity in the perturbation process is undermined and the quality of the generated pseudo-label is limited. On the other hand, increasing the number of style images might not be a good idea as well since it does not necessarily improve the performance significantly when the performance starts to be saturated despite of the increase in the computational cost.

\section{Conclusion}


We proposed a Bidirectional Style-induced Domain Adaptation (BiSIDA) framework that optimizes a segmentation model via target-guided supervised learning and source-guided unsupervised learning. With the employment of our continuous style-induce image generator, we show the effectiveness of learning from the unlabeled target dataset by providing high-dimensional perturbations for consistency regularization. Furthermore, we also reveal that the alignment between the source and the target domain from both directions without requiring adversarial training is achievable.



\clearpage
{\small
\bibliographystyle{ieee_fullname}
\bibliography{uda}
}

\end{document}